\documentclass[11pt]{article}

\usepackage[]{acl}
\usepackage{times}
\usepackage{latexsym}
\usepackage[T1]{fontenc}
\usepackage[utf8]{inputenc}
\usepackage{microtype}
\usepackage{graphicx}
\usepackage{amsmath}
\usepackage{amssymb}
\usepackage{xcolor}
\usepackage{pifont}
\usepackage{booktabs}

\title{Ablating Archetypes: The Stability of Archetypal SAEs is an Artifact of Initialization and Metric Design}

\author{
\textbf{Micha{\l}~Brzozowski}$^{1,\dagger}$ \and
\textbf{Neo~Christopher~Chung}$^{1,2}$ \\
[0.6em]
$^{1}$Samsung AI Center, Warsaw, Poland \quad
$^{2}$University of Warsaw, Poland \\
$^{\dagger}$Corresponding author: \texttt{m.brzozowsk3@samsung.com}
}

\begin{document}
\maketitle

\begin{abstract}
Dictionary learning with sparse autoencoders (SAEs) produces overcomplete bases from neural network activations that are often  interpretable and reduces polysemanticity. However, features from SAEs vary substantially across random seeds -- a
problem known as instability. Archetypal SAEs \citep{fel2025archetypal} were
proposed as a general dictionary-learning intervention for more reliable concept
extraction, and report more stable dictionaries at the end of training. We demonstrate that the stability claimed by archetypal SAEs is a result of setting identical initialization across multiple runs. Through our analyses, we attempt to clarify two distinct notions in mechanistic interpretability that may be ambiguously used: \textbf{stability} is agreement between two independently trained models, whereas \textbf{stabilization} is the convergence of independently initialized runs toward a common solution. This distinction is critical for mechanistic interpretability of natural language processing (NLP), where
feature stability is increasingly used as evidence that SAE features are
reusable units of analysis. Experiments from archetypal SAEs share a deterministic
k-means decoder initialization, setting inter-run dictionary distance to zero
before training begins. When this initialization is removed, the
archetypal constraint provides no stabilization advantage in our setting. We further identify
a preprocessing-dependent cosine geometry issue that complicates interpretation of endpoint stability metrics. Overall, our study supports the value of
studying SAEs within the larger dictionary-learning tradition while showing that
stability claims require trajectory diagnostics and initialization ablations.
\end{abstract}

\section{Introduction}

\begin{figure*}[t]
  \centering
  \includegraphics[width=\textwidth]{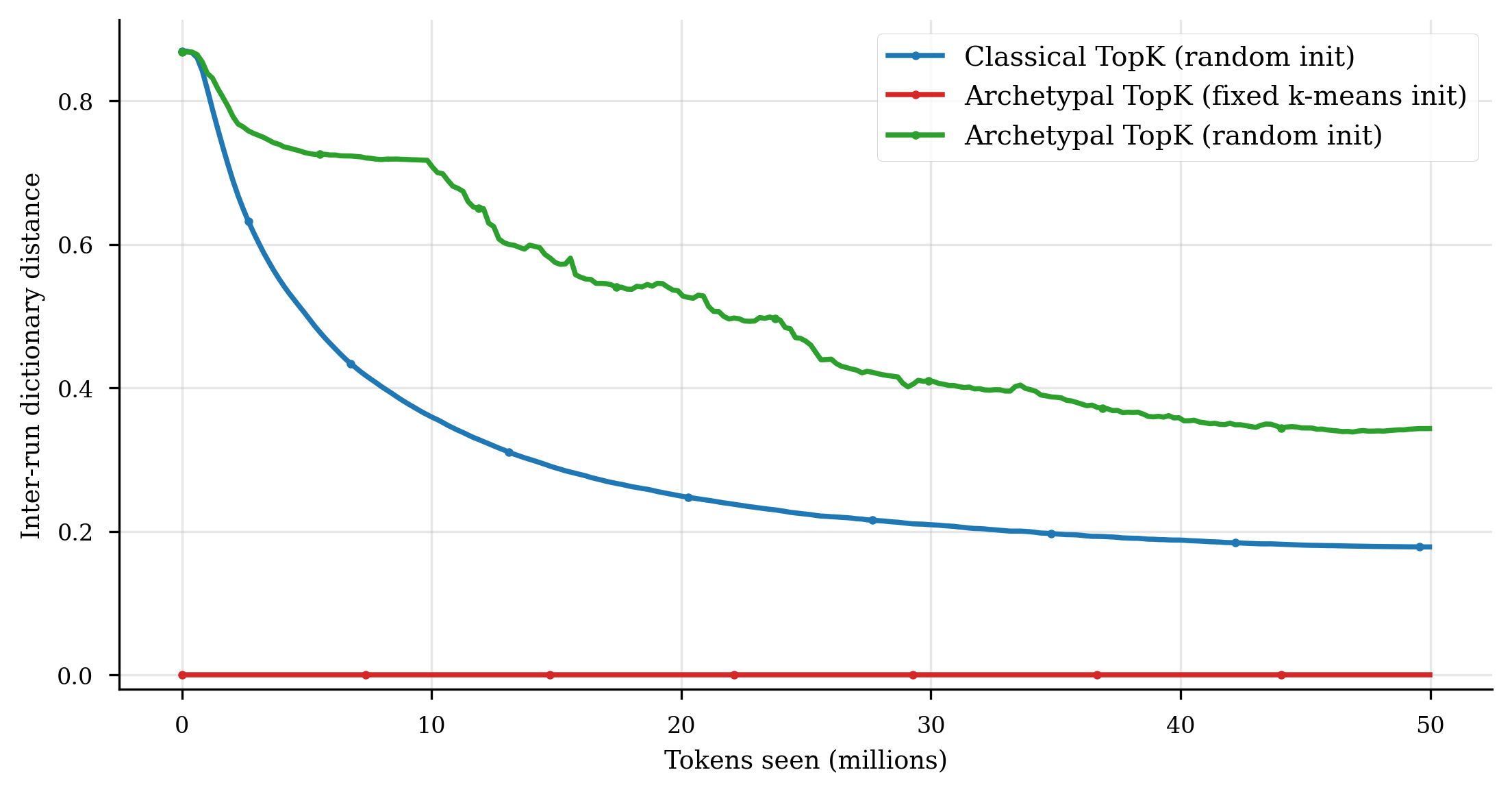}
  \caption{Inter-run dictionary distance (mean minimum cosine distance) as a
  function of tokens seen, for three conditions trained on Pythia-160m layer~6
  residual stream activations with centering applied.
  \textbf{Classical TopK} (random init, blue) starts high and converges
  monotonically: genuine stabilization driven by the data distribution.
  \textbf{Archetypal TopK} (fixed k-means init, red) is flat at zero
  throughout: both seeds start from identical dictionaries ($D = \mathbf{C}$,
  $W = I$), so no convergence is tested.
  \textbf{Archetypal TopK} (random init, green) also starts high and decreases,
  but converges at roughly half the rate of classical TopK, ending at
  ${\approx}0.35$ vs.\ ${\approx}0.17$ after 50M tokens.
  The convex-hull constraint does not accelerate stabilization in this setting;
  endpoint stability largely reflects the shared initialization.}
  \label{fig:trajectory}
\end{figure*}

Sparse autoencoders have become a standard tool in mechanistic interpretability,
especially for language models, by decomposing neural network activations into a
sparse linear combination of interpretable dictionary atoms
\citep{huben2023sparse, bricken2023monosemanticity}.
A well-known practical challenge is \emph{instability}: two SAEs trained on
identical data with different random seeds learn substantially different
dictionaries, making it unclear which features are meaningful rather than
accidental.
Recent work has treated this as a serious problem, with \citet{song2025position}
and \citet{paulo2026sparse} both independently identifying and measuring it.

Archetypal SAEs \citep{fel2025archetypal} are part of a broader and valuable
effort to place SAEs within the older dictionary-learning tradition, alongside
methods such as k-means, PCA, NMF, and archetypal analysis
\citep{fel2023holistic,cutler1994archetypal}.
This perspective is especially useful for mechanistic interpretability,
where SAEs are sometimes treated as the de facto standard for unsupervised feature
discovery rather than as one point in a larger design space. Archetypal SAEs address instability by constraining dictionary atoms to lie
within the convex hull of a set of data points (the ``archetypes''), initialized
via k-means clustering.
The authors demonstrate improved endpoint stability and attribute it to the
archetypal constraint.
We argue this attribution rests on an experimental design that does not isolate
the archetypal constraint from its deterministic initialization.

First, the conceptual issue is a conflation of \textbf{stability} with \textbf{stabilization}. Stability, which also called consistency \citep{song2025position}, is achieved by having two independently trained instances agree at the end of training. On the other hand, stabilization -- a stronger claim than stability -- requires two different initialization converge over the course of training. The former
can be achieved trivially by fixing the starting point; the latter cannot.
Archetypal SAEs report the former while never testing the latter, because both
seeds start from the same k-means centroids. Inter-run distance is zero before
a single gradient step is taken, and ``stability'' at the end is simply that
initialization being preserved. Figure~\ref{fig:trajectory} makes this
distinction visible by plotting the full inter-run trajectory rather than only
the endpoint.

Second, a more subtle issue concerns the geometry induced by preprocessing
and cosine distance. Cosine distance, the standard stability metric, can be
deflated when dictionary atoms cluster far from the origin, while element-wise
standardization defines the archetypal hull in a coordinate-dependent
transformed space. This preprocessing ambiguity complicates interpretation of
endpoint stability metrics. Our main conclusion does not depend on this metric
critique: even under the authors' preferred geometry, a stability comparison
that shares the entire decoder initialization across archetypal runs cannot
establish stabilization.

The stakes are not merely methodological. Findings of dictionary learning with SAEs
\footnote{Generally speaking, a specific combination of neurons encodes a specific concept, that a circuit
implements a specific computation} rest on the assumption that the learned
decomposition accurately reflects the underlying mechanisms of the model, rather than artifacts of the training run.
If two SAEs trained on the same data with different random seeds learn
substantially different dictionaries, it is unclear which atoms are genuine
features and which are noise.
End-of-training stability has therefore been proposed as a proxy for feature
validity: an atom that reappears across seeds is more likely to correspond to
real computational structure, not to a local optimum specific to one run.
We argue that this proxy is only meaningful, if stability is evidence of convergence.
A model that is stable by construction (because its initialization is fixed
rather than because its training dynamics converge) offers no such evidence.
The distinction matters for every downstream claim that uses stability to
validate interpretability findings.

We make three contributions:
\begin{itemize}
\item \textbf{Stabilization diagnostic}: plotting inter-run dictionary
distance as a function of training step. A flat trajectory at zero is not
stability---it is the absence of a test. When the test is administered,
classical TopK stabilizes faster than the archetypal variant.
\item \textbf{Initialization ablation}: two controlled conditions (archetypal
with the convex-hull projection disabled; classical TopK initialized from the
same k-means centroids) that isolate the contribution of the archetypal
mechanism from that of shared initialization.
\item \textbf{Metric analysis}: exposing the uncentered-cosine confound in
language-model settings and discussing how preprocessing choices affect the geometry in
which the archetypal hull and cosine stability metric are defined.
\end{itemize}

\section{Related Work}
\label{sec:related}

\paragraph{Concept extraction as dictionary learning.}
Concept-based interpretability has long studied unsupervised feature discovery
through classical dictionary-learning methods, including k-means, PCA, NMF, and
archetypal analysis. \citet{fel2023holistic} give a particularly useful
unifying framework: concept extraction is cast as dictionary learning, while
concept importance estimation is cast as attribution in the learned concept
space. We adopt this broad view. SAEs should not be treated as the only possible
unit of analysis for neural representations, including in NLP; rather, they are
one scalable neural dictionary-learning method among many. Our contribution is
not to reject this program, but to examine whether one proposed intervention in
that program---the archetypal constraint---actually supports the stability claim
made for it.

\paragraph{SAE stability.}
The instability of sparse autoencoders was independently identified and measured
by \citet{paulo2026sparse} and \citet{song2025position}.
The basic measurement idea is older than these papers: the first public SAE
interpretability report used mean maximum cosine similarity (MMCS) to compare
learned dictionaries to ground-truth features and to compare dictionaries of
different sizes \citep{sharkey2022taking}.
Hungarian matching and PW-MCC are stricter or differently normalized variants of
the same general recipe: compare dictionary atoms by pairwise similarity while
accounting for permutation invariance.
Paulo and Belrose use a cosine Hungarian loss to document that different
training runs produce substantially different dictionaries; Song et al.\ frame
the same phenomenon as a feature consistency problem and argue it undermines the
validity of circuit-level interpretability claims.
Both works vary initialization across seeds, the convention we adopt here.
Our contribution is orthogonal: we are not measuring how unstable classical SAEs
are, but asking whether the claimed fix (the archetypal constraint) actually
stabilizes anything.
In concurrent work, \citet{brzozowski2026alignedtrainingparameterfreemethod} address instability by reparameterizing
the encoder to enforce a unit inner product between each encoder row and its
corresponding decoder column, achieving Pareto improvements on reconstruction,
dead features, and cross-seed stability without additional hyperparameters.
That work demonstrates that constraining the encoder--decoder geometry is a viable
route to stabilization; ours shows that the archetypal constraint---which operates
on the decoder alone---does not constitute one.
\citet{song2025position} also report feature-consistency curves over training;
we build on this temporal perspective to separate endpoint agreement from
stabilization, i.e.\ convergence from genuinely different initial conditions.
This distinction is the central issue in the archetypal SAE setting, where
shared decoder initialization can make endpoint agreement look like stability
without testing convergence at all.

\paragraph{Archetypal SAEs.}
The archetypal constraint draws on archetypal analysis \citep{cutler1994archetypal},
a classical technique that represents data as convex combinations of extreme
points (archetypes) lying on the boundary of the data cloud.
\citet{fel2025archetypal} is the work we directly examine.
Our critique is specifically about the stability claim; the archetypal
construction has genuine virtues independent of that claim.
Constraining atoms to the convex hull of a fixed reference set provides a
principled geometric interpretation (atoms are literally combinations of
representative data points) and enforces bounded atom norms as a regularizer.
These properties may be valuable for visualization, for controlling atom scale,
or for aligning learned features with semantically meaningful prototypes.
We leave open whether the archetypal constraint provides benefits beyond
stability, and make no claim about the broader value of the method.

\paragraph{Decoder centering and initialization.}
The practice of initializing the decoder bias at the geometric median of the
training distribution, as used by \citet{bricken2023monosemanticity} and
\citet{gao2025scaling}, is conceptually related to our centering correction.
Both interventions ground the dictionary in the data distribution before
training begins.
The key distinction is the reason.
Prior work motivates centering as a convergence aid: starting near the data mean
reduces the distance atoms must travel early in training.
Our argument is different: centering is required for the cosine stability metric
to be a valid measurement instrument.
Without it, any method whose atoms remain near the data mean will appear stable
by cosine distance, regardless of the underlying learning dynamics.
The same preprocessing step serves two purposes, and the measurement-validity
purpose is the one that has not previously been stated.

\section{Methods and Metrics}
\label{sec:methods}

Let $X \in \mathbb{R}^{n \times d}$ denote a matrix of activations. A sparse
autoencoder learns an encoder $f_\theta : \mathbb{R}^d \to \mathbb{R}^m$ and
a decoder dictionary $D \in \mathbb{R}^{m \times d}$ such that
\[
  \hat{x} = zD, \qquad z = f_\theta(x),
\]
with $z$ constrained or regularized to be sparse. A TopK SAE enforces sparsity
by retaining only the $k$ largest activations in $z$ and setting the rest to
zero. In a classical TopK SAE, the decoder dictionary $D$ is a free learned
parameter, usually initialized randomly.

Archetypal SAEs replace this free dictionary with a convex-hull parameterization.
Let $\mathbf{C} \in \mathbb{R}^{r \times d}$ be a fixed set of candidate
archetypes, typically k-means centroids of the activation dataset. The
archetypal dictionary is
\[
  D = W\mathbf{C},
  \qquad W_i \in \Delta^{r-1},
\]
where each row $W_i$ lies on the probability simplex. Thus each dictionary atom
is a convex combination of the candidate points. The scalable implementation of
\citet{fel2025archetypal} initializes $W$ as the identity (or the corresponding
rectangular identity when $m \leq r$), so when the same $\mathbf{C}$ is shared
across seeds the decoder dictionary is identical at step~0.

The relaxed archetypal SAE (RA-SAE) adds a bounded residual term,
\[
  D = W\mathbf{C} + \Lambda,
  \qquad W_i \in \Delta^{r-1},
  \qquad \|\Lambda_i\|_2 \leq \delta.
\]
The hyperparameter $\delta$ controls how far each atom may move away from the
convex hull of $\mathbf{C}$. When $\delta = 0$, the atoms remain exactly inside
the hull. Larger $\delta$ permits more reconstruction flexibility while keeping
the dictionary close to the archetypal reference set.

\paragraph{Stability metrics.}
Given two trained dictionaries $D^{(1)}, D^{(2)} \in \mathbb{R}^{m \times d}$,
the standard endpoint stability metric is the optimal cosine matching between
their atoms. Let
\[
  c(i,j) =
  1 - \frac{\langle D^{(1)}_i, D^{(2)}_j\rangle}
           {\|D^{(1)}_i\|_2\|D^{(2)}_j\|_2}
\]
be cosine distance. The normalized cosine Hungarian distance is
\begin{equation}
  H(D^{(1)}, D^{(2)})
  =
  \frac{1}{m}
  \min_{\pi \in S_m}
  \sum_{i=1}^{m} c(i,\pi(i)),
  \label{eq:hungarian}
\end{equation}
where $S_m$ is the set of permutations of the $m$ atoms. Lower values indicate
more similar dictionaries. This is the metric we report for endpoint stability
in Table~\ref{tab:results}.

For training trajectories, computing the Hungarian assignment at every logging
step is unnecessarily expensive. We therefore use the mean minimum cosine
distance
\[
  M(D^{(1)}, D^{(2)})
  =
  \frac{1}{m}\sum_{i=1}^{m}\min_j c(i,j).
\]
This drops the bijection constraint and asks how close each atom in one
dictionary is to its nearest neighbor in the other. \citet{paulo2026sparse}
argue that Hungarian matching is more principled, but empirically find that
maximum-cosine nearest-neighbor scores closely track the Hungarian scores
because most atoms are matched to their nearest neighbors. We use $M$ only as a
linear-time trajectory diagnostic; endpoint numbers use the Hungarian metric.

\section{Clarifying Confusion}
\label{sec:confounds}
The concept of SAE stability is less well-defined in the literature than it
might appear, and the evidence for archetypal stability draws on two distinct
measurement pathologies. The first is an inconsistency in what randomness is
varied across seeds; the second is a geometric pathology of the cosine metric
on uncentered activations. Together they inflate reported stability scores
for any method whose dictionary atoms remain near the data mean.

\subsection{Inconsistent sources of randomness}

Randomness can enter SAE training in two distinct ways: \emph{parameter
initialization}, which determines the starting point of optimization, and
\emph{batch order}, which affects the trajectory of stochastic gradient descent.
Conflating these leads to incomparable experimental results.

Table~\ref{tab:consistency} organizes the assumptions made by recent papers
on SAE stability. \citet{paulo2026sparse} and \citet{song2025position} both
vary initialization across seeds while fixing batch order, whereas
\citet{fel2025archetypal} vary batch order but fix initialization: k-means
always produces the same centroids given the same data and seed.
This inconsistency means that archetypal SAEs enjoy a structural advantage in
every stability comparison against classical SAEs. They are never exposed to
the primary source of instability.

\begin{table}[h]
\centering
\small
\begin{tabular}{lcc}
\toprule
\textbf{Paper} & \textbf{Rand.\ init} & \textbf{Rand.\ batches} \\
\midrule
\citet{paulo2026sparse} & \textcolor{green}{\ding{52}} & \textcolor{red}{\ding{56}} \\
\citet{song2025position}    & \textcolor{green}{\ding{52}} & \textcolor{red}{\ding{56}} \\
\citet{fel2025archetypal}   & \textcolor{red}{\ding{56}} (fixed)\textsuperscript{$\dagger$} & \textcolor{green}{\ding{52}} \\
\bottomrule
\end{tabular}
\caption{Sources of randomness across recent papers on SAE stability.
Archetypal SAEs fix initialization by design, giving them an unacknowledged
advantage over classical SAEs in all published comparisons.
\textsuperscript{$\dagger$}Archetypal SAEs do vary \emph{encoder} weights across seeds; we
classify initialization as fixed because the stability metric is computed on the
\emph{decoder} dictionary, which is deterministic at step~0: $D = W \mathbf{C}$
where $\mathbf{C}$ is the fixed centroid matrix and $W$ is initialized to the
identity. Encoder randomness reaches the dictionary only through gradients
during training.}
\label{tab:consistency}
\end{table}

Recent literature contains a direct conflict about whether TopK SAEs are more
or less stable than vanilla SAEs with L1 regularization:
\citet{paulo2026sparse} find TopK SAEs more seed-dependent than ReLU/L1 SAEs,
whereas \citet{song2025position} find TopK SAEs more consistent than Standard
SAEs in their selected settings.
This disagreement was also raised in discussion of \citet{paulo2026sparse} \footnote{https://openreview.net/forum?id=EjInprGpk9},
where the authors noted that several recent works report stable L1-trained SAEs
and suggested that differences in dead-latent rates may explain part of the
discrepancy.
We do not adjudicate that architecture-level question here; our point is that
stability comparisons are sensitive to the experimental definition of
``independent runs,'' including which sources of randomness are varied and which
features are counted.

\subsection{The uncentered-cosine confound}
\label{sec:cosine-confound}

Cosine distance is a meaningful stability metric only when the objects being
compared are centered near the origin. In LLM residual streams, activations
can have a non-negligible mean component: the distribution sits away from zero.
Under k-means clustering of uncentered activations,
every resulting centroid inherits this mean offset. All centroids, and therefore
all initial dictionary atoms of an archetypal SAE, lie inside a narrow cone
around $\mu$.

The cosine distance between two vectors in such a cone is dominated by $\mu$,
not by the directions that distinguish them. This is the \emph{galaxy-far-away}
effect: all stars in a distant galaxy have nearly identical angles as seen from
Earth, even though they may be far apart in absolute terms.
\citet{li2025geometry} independently characterize this large-scale anisotropy
of SAE feature clouds and use the same galaxy metaphor to describe it.

\begin{figure*}[t]
\centering
\includegraphics[width=\textwidth]{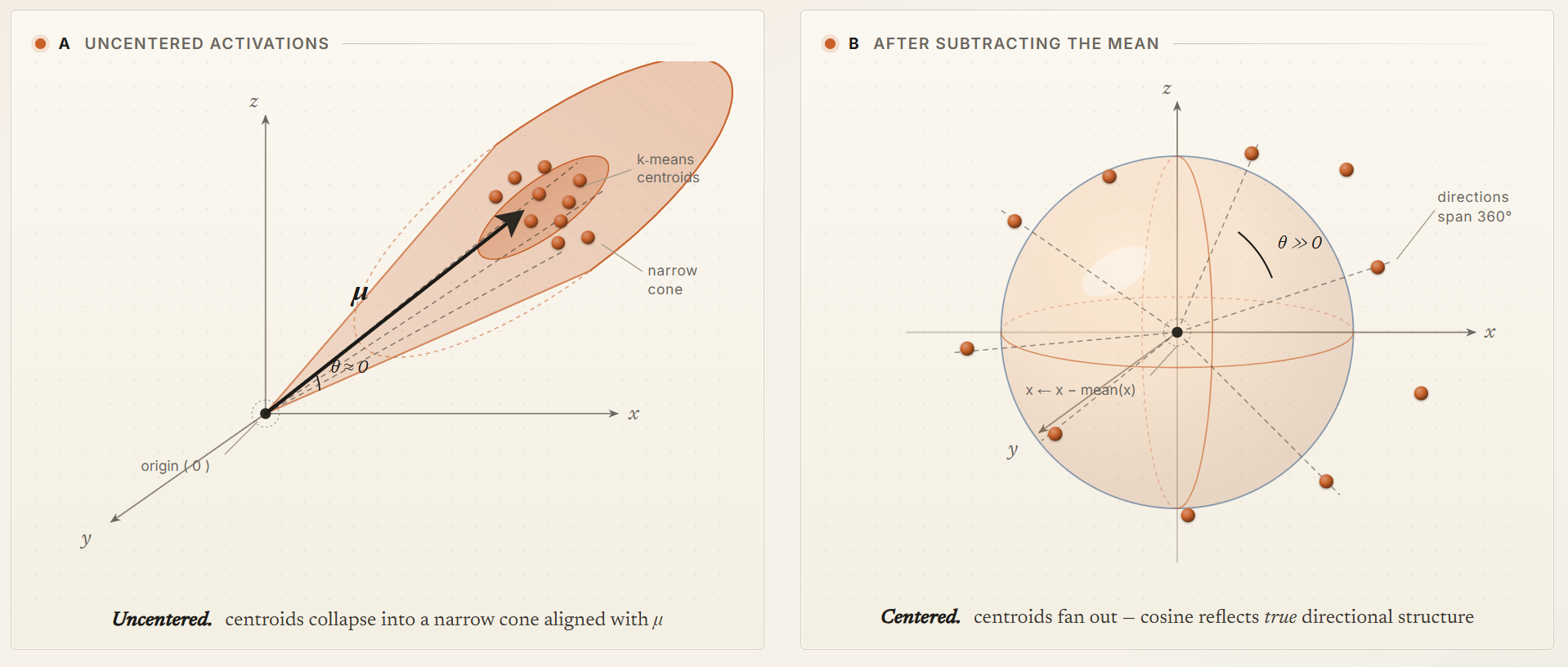}
\caption{\textbf{The galaxy-far-away effect.} Cosine similarity measures angles
from the \emph{origin}, not distances between points. \textbf{(a)} When
activations have a large mean $\mu$, all k\mbox{-}means centroids cluster
far from the origin in a narrow cone. Any two atoms in this cone appear highly
similar by cosine distance: not because they encode similar features, but
because the mean offset $\mu$ dominates all directions. \textbf{(b)} Centering
(subtracting $\mu$) scatters the centroids around the origin; cosine distance
now reflects genuine directional differences.}
\label{fig:galaxy}
\end{figure*}

Figure~\ref{fig:galaxy} illustrates the geometry. Panel~(a) shows k-means
centroids clustered near the mean vector $\mu$; the angular spread seen from
the origin is tiny, so any two atoms score as highly similar regardless of
what they encode. Panel~(b) shows the same atoms after centering: the mean offset
is removed, atoms scatter around the origin, and cosine distance now
reflects genuine directional differences.

This is a measurement artifact, not a property of the learned dictionary.
Any method that constrains its atoms to lie near the data mean (whether
through the archetypal convex hull, through fixed k-means initialization,
or simply through an objective that does not push atoms away from $\mu$) will
appear stable by cosine distance on uncentered activations. Centering the
activations before running k-means, and subtracting the activation mean from
the decoder bias during training, eliminates this pathology. We apply this correction to both experimental settings
(\S\ref{sec:trajectory}, \S\ref{sec:ablation}).

A natural question is whether \citet{fel2025archetypal} apply such a correction.
The paper states that activations are element-wise standardized, which would
eliminate the mean offset and thereby fix the cosine metric.
However, neither the public Colab notebook nor the \texttt{Overcomplete} library
applies any centering or standardization before k-means or during training,
so the reported stability numbers are computed without this correction.
Moreover, element-wise standardization---dividing each coordinate by its
estimated standard deviation---is not geometrically motivated in LLM settings:
it privileges the standard basis of the weight-storage coordinate system, which
is arbitrary for residual stream activations.
Centering alone is sufficient to make the cosine metric valid; per-coordinate
rescaling by standard deviation adds an unjustified coordinate-dependent
assumption.
See Appendix~\ref{sec:code-audit} for the detailed code audit supporting these claims.

\subsection{Empirical verification}
\label{sec:cone-empirical}

Table~\ref{tab:cone} reports the key statistics for both experimental settings,
using the centroid sets used in our stability experiments.

\begin{table}[h]
\centering
\resizebox{\columnwidth}{!}{%
\begin{tabular}{lrr}
\toprule
 & \multicolumn{2}{c}{Mean pairwise cosine sim} \\
\cmidrule(lr){2-3}
 & Uncentered & Centered \\
\midrule
LLM (Pythia-160m, L6) & 0.362 & 0.002 \\
Vision (DinoV2)       & 0.298 & $-$0.001 \\
\bottomrule
\end{tabular}}
\caption{Mean pairwise cosine similarity among k-means centroids before and
after centering. Centering removes a systematic positive bias of 0.30--0.36 in
cosine similarity in both settings.}
\label{tab:cone}
\end{table}

The key observation is the before/after change in centroid cosine similarity.
Before centering, mean pairwise cosine similarity among centroids is 0.362
(LLM) and 0.298 (vision), which is not negligible.
After centering, both collapse to approximately zero, consistent with
centroids that are isotropically distributed around the origin.
This positive bias directly inflates reported stability: cosine
\emph{distance} is 1 minus cosine similarity, so any two centroids appear
systematically closer by 0.30--0.36 units when activations are not centered.
A stability method that keeps atoms near the uncentered data mean will benefit
from this inflation regardless of whether it has learned anything useful.

Notably, DinoV2 vision centroids show the same positive cosine bias before
centering (0.298), confirming that the same confound applies in the vision
setting.
We therefore apply the centering correction to both experimental settings
in \S\ref{sec:trajectory} and \S\ref{sec:ablation}.

\section{Stabilization, not Stability}
\label{sec:trajectory}

End-of-training stability and stabilization are not the same thing. A model
is \emph{stable} if two independently trained instances agree at the end.
A model \emph{stabilizes} if those instances, started from genuinely different
points, converge over the course of training. The former can be achieved
trivially by fixing the starting point; the latter cannot. We argue
stabilization is the right desideratum---it is evidence that the data
distribution is constraining the learned representation, rather than an artifact
of the experimental protocol.

Any method that reduces variance in the starting point of optimization will
look stable by an end-of-training measure, regardless of whether the underlying
learning dynamics converge. This is the identification problem for archetypal
SAE stability claims.
The decoder dictionary is $D = W\mathbf{C}$, where $\mathbf{C}$ is the fixed
centroid matrix (computed once, shared across all seeds) and $W$ is initialized
to the identity. At step~0, before any gradient is applied, the two seeds
produce \emph{identical} decoder dictionaries: $D = \mathbf{C}$.
Inter-run decoder distance is not merely small at initialization---it is
exactly zero by construction.
End-of-training stability then reflects this identity being approximately
preserved, not any convergence that happened during training.

The diagnostic we propose is to track inter-run distance throughout training,
not just at the end. This temporal view is related to the training-dynamics
analysis of \citet{song2025position}, but the conceptual target here is
different: \emph{stabilization}, not stability. If two independently trained
SAEs are genuinely converging to the same solution, their dictionary distance
should \emph{decrease} over the course of training, starting high (different
initializations) and falling as the shared structure of the data constrains the
solution. A method that merely preserves a shared initialization shows endpoint
agreement without stabilization. In the limiting case relevant here, the runs
begin identical: the two runs were never meaningfully separated to begin with.

Figure~\ref{fig:trajectory} shows this trajectory for the conditions we
study, measured as mean minimum cosine distance between paired dictionaries
at regular training intervals (defined in \S\ref{sec:methods}).%

\textbf{Classical TopK} with random initialization starts with high inter-run
distance and decreases monotonically over training: genuine stabilization.
The two seeds begin from unrelated random points in a high-dimensional parameter
space and converge as training progresses.
This is what stabilization looks like: the data distribution has a basin of
attraction, and both runs find it independently.

\textbf{Archetypal TopK} with fixed k-means initialization is flat at zero from
step~0.
As established above, both seeds produce identical decoder dictionaries at
initialization ($D = \mathbf{C}$), and this identity is largely preserved
throughout training.
The relaxation term $\delta$ bounds how far atoms can drift from the centroid
set, so inter-run distance at convergence is also bounded by $2\delta$, small
by design.
This is not convergence. The two runs were never apart.

\textbf{Archetypal TopK} with random initialization provides the
decisive test of whether the convex-hull constraint itself exerts a convergence
force, independent of the shared starting point.
Here the decoder initialization is randomized: the convex combination weights
$W$ and relaxation offsets are drawn randomly per seed, while the centroid
set $\mathbf{C}$ remains fixed and centered.
Both seeds therefore start from genuinely different decoder dictionaries
(inter-run distance $\approx 0.88$ at step~0), and the archetypal constraint
is the only structural difference from the classical condition.

The trajectory does decrease (the constraint does exert some convergence
force) but at roughly half the rate of unconstrained classical TopK.
After 50 million tokens, the archetypal random-init condition reaches a
distance of $\approx 0.35$, compared to $\approx 0.17$ for classical TopK.
Rather than accelerating stabilization, the convex-hull constraint slows it.
This reversal is decisive: atoms that are free to move wherever the data
pulls them converge to shared solutions faster than atoms constrained to the
interior of the centroid hull.
In this setting, the archetypal mechanism does not enhance stabilization; it
impedes it. The endpoint stability reported by \citet{fel2025archetypal}
therefore cannot be interpreted as evidence that independently initialized
archetypal dictionaries converge to a shared solution.

\section{Ablating Archetypes}
\label{sec:ablation}

We now isolate the mechanism behind the stability claim with two ablations,
using the minimal working example provided by the authors of archetypal SAEs as
a public Google Colab
notebook.\footnote{\url{https://colab.research.google.com/drive/1TmAtUhIdFGSMlDhKr2ndXGR8GU4R4aTq}}
This setup uses DinoV2 patch embeddings extracted from a single image class
(rabbits).
We additionally replicate all experiments on residual stream activations from
layer~6 (the middle layer) of Pythia-160m \citep{biderman2023pythia}, using
WikiText-103 as the text source.
In both settings, stability is measured as the normalized cosine Hungarian
loss between the dictionaries of two independently trained SAEs
(Eq.~\ref{eq:hungarian}); a lower score indicates more stable (similar)
dictionaries across seeds.
Unlike the original notebook, we fix \texttt{torch.manual\_seed} before each
training run to ensure that any difference between conditions is attributable
to initialization rather than encoder weight randomness.
Our code is available at \url{https://github.com/MichalBrzozowski91/ablating-archetypes}.
The repository provides two notebooks: one replicating the original Colab's
preprocessing exactly (uncentered, for direct comparison), and one with the
centering correction applied (matching Table~\ref{tab:results}).
See the repository README for details.

Both settings center activations before running k-means and subtract the
activation mean from the decoder bias before training.
As documented in \S\ref{sec:cone-empirical}, this correction is necessary for
the cosine stability metric to reflect genuine directional differences rather
than proximity to the uncentered data mean.

\subsection{Ablation 1: Disabling the Convex-Hull Projection}

The defining feature of archetypal SAEs is the projection that keeps dictionary
atoms within the convex hull of the data points.
If this constraint is responsible for stability, disabling it should degrade
stability toward the randomly initialized classical SAE baseline.
We test this by training an archetypal SAE with the convex-hull projection
disabled, retaining the k-means initialization but removing the convex-hull
enforcement during training.
Without the projection, both the convex-hull constraint and the $\delta$ bound
on the residual term $\Lambda$ are disabled: $W$ is frozen, the norm constraint
$\|\Lambda_i\|_2 \leq \delta$ is lifted, and $\Lambda$ becomes a free, unconstrained
learnable dictionary matrix initialized directly to $\mathbf{C}$.
It is this combination --- removing the projection \emph{and} releasing the
residual bound --- that makes the ablated model structurally identical to a
classical TopK SAE whose decoder has been initialized from the k-means centroids,
which is exactly Condition 4 (\S\ref{sec:ablation2}).
The numerical identity of Conditions 3 and 4 in Table~\ref{tab:results} (identical
Hungarian distance and R$^2$ under consistent seeding) follows by construction,
not coincidence: they are running the same forward and backward pass, implemented
two ways.

As shown in Table~\ref{tab:results}, the ablated model is \emph{more} stable
than the full archetypal SAE in both experimental settings
(vision: 0.068 vs.\ 0.243; Pythia-160m: 0.112 vs.\ 0.170).

The R$^2$ gap shows that the convex-hull projection hurts reconstruction.
This is the default expectation under any constraint: restricting the feasible
set can only reduce reconstruction capacity unless the constraint happens to be
a well-suited inductive bias for the data.
Whether the archetypal parameterization constitutes such a bias is an open
question, and the burden of proof lies with its proponents.
Our results are consistent with the reduced reconstruction performance
acknowledged by \citet{fel2025archetypal} themselves.
Crucially, the projection is not necessary for the observed stability either,
so it incurs a reconstruction cost with no stabilization benefit in our setting.

\subsection{Ablation 2: Fixing the Initialization of a Classical SAE}
\label{sec:ablation2}

If stability comes from initialization rather than the archetypal constraint,
then a classical TopK SAE initialized from the same k-means centroids should
achieve the same stability as an archetypal SAE.
We construct this condition by initializing the dictionary of a classical TopK
SAE directly from the first $K$ k-means centroids, with no archetypal constraint
applied at any point during training.

The results support the initialization account.
In both settings, the classical SAE with fixed initialization matches the ablated
archetypal SAE exactly (vision: 0.068; Pythia-160m: 0.112) and substantially
outperforms both the randomly initialized classical SAE (vision: 0.433;
Pythia-160m: 0.200) and the full archetypal SAE with its convex-hull projection
intact (vision: 0.243; Pythia-160m: 0.170).

This result has a natural interpretation: k-means initialization places dictionary
atoms near dense regions of the activation space, reducing the effective
distance each atom must travel during optimization and thereby making training
less sensitive to the stochastic batch order.
This is conceptually related to the decoder bias centering technique proposed
in \citet{bricken2023monosemanticity} and used in \citet{gao2025scaling}, where
subtracting the geometric median of activations from the decoder bias serves
a similar role of grounding the dictionary in the data distribution before
training begins.

\begin{table}[t]
\centering
\resizebox{\columnwidth}{!}{%
\begin{tabular}{lcccc}
\toprule
 & \multicolumn{2}{c}{\textbf{Vision (DinoV2)}} & \multicolumn{2}{c}{\textbf{Language Model (Pythia-160m)}} \\
\cmidrule(lr){2-3} \cmidrule(lr){4-5}
\textbf{Condition} & \textbf{Hung.\,$\downarrow$} & \textbf{R$^2$\,$\uparrow$} & \textbf{Hung.\,$\downarrow$} & \textbf{R$^2$\,$\uparrow$} \\
\midrule
Classic SAE (random init)  & 0.433 & 0.573 & 0.200 & 0.937 \\
Archetypal SAE             & 0.243 & 0.542 & 0.170 & 0.758 \\
Archetypal SAE (ablated)   & 0.068 & 0.570 & 0.112 & 0.914 \\
Classic SAE + fixed init   & 0.068 & 0.570 & 0.112 & 0.914 \\
\bottomrule
\end{tabular}}
\caption{Normalized cosine Hungarian distance (lower = more stable) and R$^2$
reconstruction score (higher = better) across two experimental settings.
Ablated and fixed-init conditions are numerically identical in both settings
under consistent seeding, confirming they are equivalent.
Notably, the full archetypal SAE achieves the \emph{worst} R$^2$ in both
settings, suggesting that the convex-hull projection constraint impedes
reconstruction quality while the observed stability follows from shared
initialization rather than the archetypal constraint.}
\label{tab:results}
\end{table}

\section*{Limitations}

Our experiments are conducted with a set of hyperparameters drawn from the reference
implementation of \citet{fel2025archetypal} (number of
concepts, sparsity level, training epochs).
We do not perform a hyperparameter sweep, and it is possible that the
archetypal constraint provides some benefit at other settings or in other
regimes.
The ablation experiments (\S\ref{sec:ablation}) follow the reference
implementation and train for 20 epochs; the stability trajectory experiment
(\S\ref{sec:trajectory}) trains for 50 million tokens.
In both cases, longer training may shift the relative ordering of conditions.
The vision experiment uses a single image class (rabbits) from the authors'
reference Colab, and the language experiment a single layer of a single model;
broader coverage would strengthen the empirical claim.

\bibliography{custom}

\appendix
\section{Hyperparameters: Stability Trajectory Experiment}
\label{sec:hyperparams}

Table~\ref{tab:hyperparams} lists the full hyperparameter configuration for the
LLM stability trajectory experiment (\S\ref{sec:trajectory}).
All values are script defaults; no sweep was performed.

\begin{table}[h]
\centering
\scriptsize
\begin{tabular}{ll}
\toprule
\textbf{Parameter} & \textbf{Value} \\
\midrule
\multicolumn{2}{l}{\textit{Model and data}} \\
Model & Pythia-160m-deduped \\
Layer & 6 (residual stream post-layer) \\
Activation dimension & 768 \\
Training dataset & Pile (monology/pile-uncopyrighted) \\
Context length & 1{,}024 tokens \\
Total training tokens & 50{,}000{,}000 \\
\midrule
\multicolumn{2}{l}{\textit{SAE architecture}} \\
Dictionary size & 4{,}096 \\
Sparsity ($k$) & 20 \\
Auxiliary loss weight ($\alpha_\text{auxk}$) & $1/32$ \\
Relaxation bound ($\delta$, archetypal) & 1.0 \\
\midrule
\multicolumn{2}{l}{\textit{Optimization}} \\
Optimizer & Adam \\
Learning rate & $3 \times 10^{-4}$ \\
SAE batch size & 2{,}048 \\
LR warmup steps & 1{,}000 \\
LR decay start & step 19{,}531 (80\% of training) \\
\midrule
\multicolumn{2}{l}{\textit{K-means initialization}} \\
Tokens collected for k-means & 500{,}000 \\
Number of centroids & 4{,}096 (= dictionary size) \\
Centering & applied before k-means \\
\midrule
\multicolumn{2}{l}{\textit{Trajectory logging}} \\
Inter-run distance metric & mean minimum cosine distance \\
Logging interval & every 100 steps \\
\bottomrule
\end{tabular}
\caption{Hyperparameters for the stability trajectory experiment (Figure~\ref{fig:trajectory}).}
\label{tab:hyperparams}
\end{table}

\section{Code Audit Notes}
\label{sec:code-audit}

We examined the public codebases of both papers whose methodology we discuss.
This appendix records the specific findings that bear on the claims in the main text.

\paragraph{Song et al.\ (\citeyear{song2025position}) — batch order is fixed across seeds.}
The repository is available at \url{https://github.com/xiangchensong/sae-feature-consistency}.
In \texttt{examples/run\_single.py} (lines 63--65), the global random state is
fixed once before the training loop begins:
{\small\begin{verbatim}
random.seed(demo_config.random_seeds[0])
t.manual_seed(demo_config.random_seeds[0])
np.random.seed(demo_config.random_seeds[0])
\end{verbatim}}
Individual SAE trainers subsequently receive different seeds (defaulting to 42,
43, and 44) via the \texttt{random\_seeds} loop, which controls weight
initialization.
The \texttt{ActivationBuffer} draws training batches via
\texttt{t.randperm()} (line 74 of
\texttt{examples/dictionary\_learning/buffer.py}), which reads from the global
PyTorch random state.
Because that state is fixed to \texttt{random\_seeds[0]} before the loop, all
runs see the same sequence of activation batches and differ only in their weight
initialization.
This is consistent with our Table~\ref{tab:consistency} classification: Song
et al.\ vary initialization across seeds and hold batch order fixed.

\paragraph{Fel et al.\ (\citeyear{fel2025archetypal}) — paper and code disagree on preprocessing; standardization is not geometrically motivated.}
The reference implementation is the public Colab notebook linked in the paper
(\url{https://colab.research.google.com/drive/1TmAtUhIdFGSMlDhKr2ndXGR8GU4R4aTq})
and the accompanying \texttt{Overcomplete} package
(\url{https://github.com/KempnerInstitute/Overcomplete}).

\textit{Paper vs.\ code discrepancy.}
The paper states (p.~7): ``The data matrix $A$ was element-wise standardized.''
Element-wise standardization---subtracting the per-coordinate mean and dividing
by the per-coordinate standard deviation---does eliminate the galaxy-far-away
effect, because subtracting the mean centers the distribution around the origin.
However, neither the Colab notebook nor the \texttt{Overcomplete} library applies
any centering or standardization before running k-means or during SAE training.
The stability numbers reported in the reference notebook are therefore computed
without the preprocessing described in the paper.

\textit{Standardization is not geometrically motivated.}
Even setting aside the paper--code discrepancy, element-wise standardization
carries an additional assumption that centering alone does not: dividing each
coordinate by its estimated standard deviation privileges the standard basis of
the ambient space.
For LLM residual stream activations, no such privileged coordinate system
exists---the basis in which weights happen to be stored is arbitrary and changes
under linear reparameterization.
Centering (subtracting the mean) is geometrically well-founded and sufficient to
make the cosine metric a valid stability instrument; the per-coordinate scaling
by standard deviation is an additional, unjustified assumption in this setting.
Our correction applies centering only, without per-coordinate rescaling.

We verify the practical effect empirically in Table~\ref{tab:cone}: before
centering, mean pairwise cosine similarity among k-means centroids is 0.362
(LLM) and 0.298 (vision); after centering it collapses to approximately zero in
both settings.
Any stability metric based on cosine distance is therefore deflated for any
method---including archetypal SAEs---whose dictionary atoms remain near $\mu$.
We apply the centering correction to both experimental settings
(\S\ref{sec:trajectory}, \S\ref{sec:ablation}).

\end{document}